\documentclass[sigconf]{acmart}
\AtBeginDocument{%
  }

\setcopyright{acmlicensed}
\copyrightyear{2024}
\acmYear{2024}
\acmDOI{XXXXXXX.XXXXXXX}

\acmJournal{TOG}
\acmVolume{37}
\acmNumber{4}
\acmArticle{111}
\acmMonth{8}

\acmSubmissionID{200}

\usepackage{bm}

\usepackage{bbding}
\usepackage{multirow} 
\usepackage{booktabs}
\usepackage{color, colortbl}
\definecolor{LightGray}{rgb}{0.95, 0.95, 0.95}
\definecolor{LightBlue}{rgb}{0.89, 0.93, 0.99}
\definecolor{LightRed}{rgb}{0.99, 0.93, 0.89}
\definecolor{LightGreen}{rgb}{0.65, 0.81, 0.55}
\definecolor{LightPurple}{rgb}{0.75, 0.80, 1.0}

\begin{document}

\title{360-GS: Layout-guided Panoramic Gaussian Splatting For Indoor Roaming}

\author{Jiayang Bai}
\email{jybai@smail.nju.edu.cn}
\orcid{0000-0001-6892-0338}
\affiliation{%
  \institution{Nanjing University}
  \country{China}
}

\author{Letian Huang}
\email{lthuangg@gmail.com}
\affiliation{%
  \institution{Nanjing University}
  \country{China}
}

\author{Jie Guo}
\email{ jieguo@nju.edu.cn}
\affiliation{%
  \institution{Nanjing University}
  \country{China}
}

\author{Wen Gong}
\email{lesliewinnie@163.com}
\affiliation{%
  \institution{Nanjing University}
  \country{China}
}

\author{Yuanqi Li}
\email{yuanqili@smail.nju.edu.cn}
\affiliation{%
  \institution{Nanjing University}
  \country{China}
}

\author{Yanwen Guo}
\email{ywguo@nju.edu.cn}
\affiliation{%
  \institution{Nanjing University}
  \country{China}
}




\renewcommand{\shortauthors}{Bai et al.}

\begin{abstract}

3D Gaussian Splatting (3D-GS) has recently attracted great attention with real-time and photo-realistic renderings. This technique typically takes perspective images as input and optimizes a set of 3D elliptical Gaussians by splatting them onto the image planes, resulting in 2D Gaussians. However, applying 3D-GS to panoramic inputs presents challenges in effectively modeling the projection onto the spherical surface of ${360^\circ}$ images using 2D Gaussians. In practical applications, input panoramas are often sparse, leading to unreliable initialization of 3D Gaussians and subsequent degradation of 3D-GS quality. In addition, due to the under-constrained geometry of texture-less planes (e.g., walls and floors), 3D-GS struggles to model these flat regions with elliptical Gaussians, resulting in significant floaters in novel views. To address these issues, we propose 360-GS, a novel $360^{\circ}$ Gaussian splatting for a limited set of panoramic inputs. Instead of splatting 3D Gaussians directly onto the spherical surface, 360-GS projects them onto the tangent plane of the unit sphere and then maps them to the spherical projections. This adaptation enables the representation of the projection using Gaussians. We guide the optimization of 360-GS by exploiting layout priors within panoramas, which are simple to obtain and contain strong structural information about the indoor scene. Our experimental results demonstrate that 360-GS allows panoramic rendering and outperforms state-of-the-art methods with fewer artifacts in novel view synthesis, thus providing immersive roaming in indoor scenarios.

\end{abstract}


\received{20 February 2007}
\received[revised]{12 March 2009}
\received[accepted]{5 June 2009}

\begin{teaserfigure}
  \begin{center}
  \includegraphics[width=\linewidth, trim={0px, 0px, 0px, 0px}, clip]{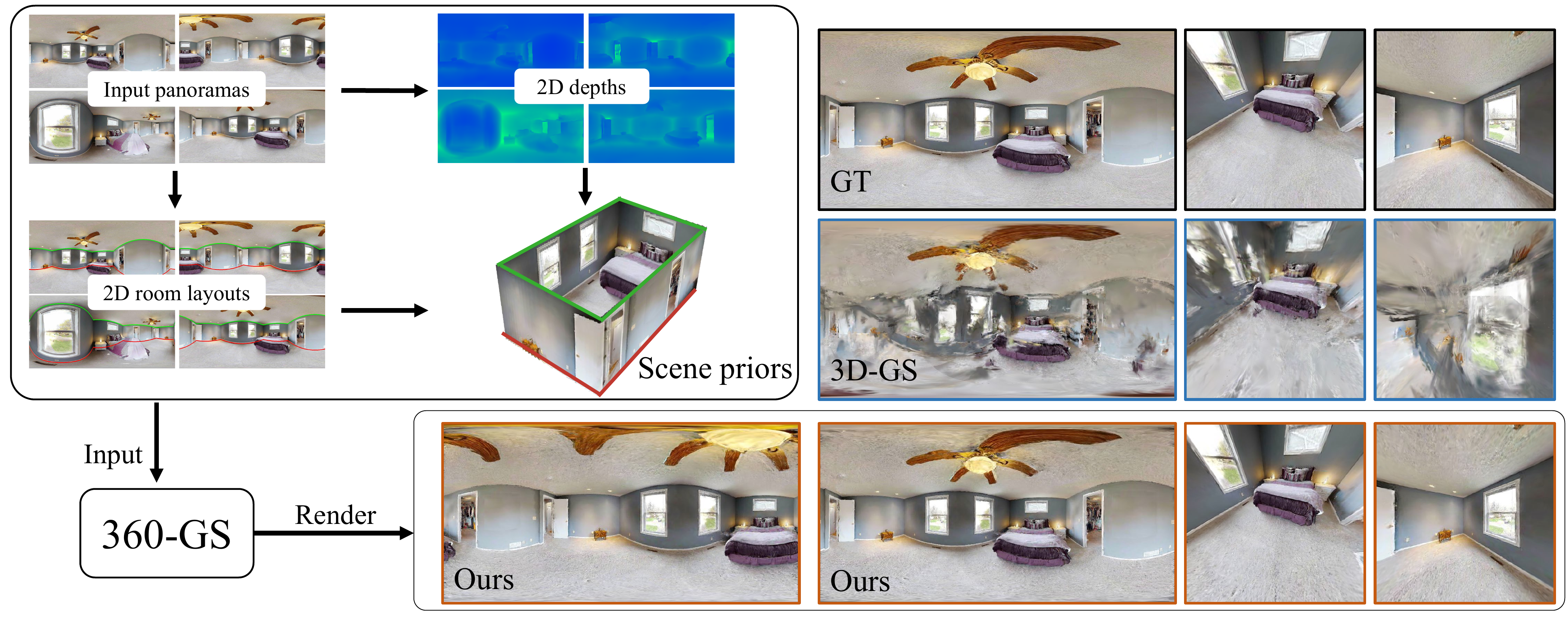}
  \end{center}
  \caption{Taking four indoor panoramas as input, our method optimizes 3D Gaussians under the guidance of scene priors including room layouts and depths. Leveraging our $360^{\circ}$ splatting algorithm, we are able to render high-quality equirectangular images from the refined 3D Gaussians. The visual comparisons illustrate that our method surpasses 3D-GS~\cite{3DGS} with accurate geometry and plausible details. }
  \label{fig:teaser}
\end{teaserfigure}
\maketitle

\section{Introduction}

With the popularity of consumer-level $360^{\circ}$ cameras, novel view synthesis from a set of panoramic images has been one of the core components of computer graphics and vision applications, including virtual and augmented reality (VR/AR). Recently, Neural Radiance Fields (NeRF)~\cite{Mildenhall2020NeRFRS, mipnerf} have attracted great attention due to their ability to produce photo-realistic renderings and become a widely used technique to synthesize novel views. Despite the massive efforts, NeRF samples dense points along the ray to render each single pixel, which is challenging for real-time rendering. Recently, point-based representation, 3D Gaussian Splatting (3D-GS)~\cite{3DGS}, has emerged as an alternative representation that achieves real-time speed with comparable rendering quality to NeRF-based methods. This promising technique enables us to roam an indoor room in real time, which has many practical applications such as free-viewpoint navigation, house touring, and virtual-reality games.

However, 3D-GS mainly focuses on perspective images. When given a set of indoor panoramas, synthesizing novel views with 3D-GS encounters several challenges. \textit{First}, splatting 3D Gaussians onto panoramic images has spatial distortion that can not be modeled with 2D Gaussians splatted onto image planes of perspective projection. Thus it is impossible to directly optimize 3D Gaussians with panoramic images. 
\textit{Second}, collecting dense panoramic views of a scene is often expensive and time-consuming~\cite{wang2023sparsenerf}. In a typical image collection process, the $360^{\circ}$ camera is usually placed at the center or in a limited set of locations in the rooms, resulting in sparse input. This scarcity of inputs significantly exacerbates the inherent ambiguity of learning 3D structure from 2D images, thus leading to unsatisfying renderings~\cite{Xiong2023SparseGSR3, FSGS}. While many works have attempted to address the few-shot task by leveraging pixel-wise information such as depth supervision~\cite{FSGS, Deng2021DepthsupervisedNF} and cross-view semantic consistency~\cite{Jain2021PuttingDiet}, the scene-level structural information within panoramas remains under-exploited. 
\textit{Third}, indoor scenes often contain many texture-less and flat regions such as walls, floors, tables, and ceilings, which are insufficient for finding cross-view correspondences. Even though 3D-GS can well fit training pixels, the geometry of these planes is inaccurate, leading to floaters above the planes in novel views. Previous works have tackled this problem through geometric regularization~\cite{Chen2022StructNeRFNR, Deng2021DepthsupervisedNF}, but most of them are built on top of NeRF.

To address the aforementioned challenges, we propose 360-GS, a novel layout-guided 3D Gaussian splatting pipeline designed for sparse panoramic images. This approach achieves real-time panoramic rendering while delivering high-quality novel views, significantly reducing undesired artifacts such as floaters, as depicted in Fig~\ref{fig:teaser}. The impressive performance is attributed to two core components of 360-GS: $360^{\circ}$ Gaussian splatting and the incorporation of room layout priors. 
$360^{\circ}$ Gaussian splatting aims to splat 3D Gaussians onto the unit sphere of panoramic inputs. We find that the spatially distorted projection is difficult to model with Gaussians. Therefore, $360^{\circ}$ Gaussian splatting algorithm decomposes the splatting into two steps: projecting 3D Gaussians onto the tangent plane and then mapping them to the spherical surface. The decomposition avoids the complicated representation of projections while maintaining real-time performance.

We further address the under-constrained problem due to few-shot inputs and texture-less planes by introducing room layout priors. 
With a full field of view, a panorama inherently contains richer global structural information than a perspective image that can be exploited for more regularization. The room layout is the most common and easy-obtained structural information for indoor scenes. From the room layout,  we derive a high-quality point cloud for the initialization of 3D Gaussians. 
Since the room layout describes the scene with flat walls, floors, and ceilings, we further enforce constraints on the positions of 3D Gaussians in these regions. The layout-guided initialization and regularization contribute to the generation of flat planes and a reduction in undesired floaters in novel views. The experiments conducted on real-world datasets have demonstrated the superiority and effectiveness of our method.

In summary, the main contributions of our paper are:
\begin{itemize}
    \item We propose 360-GS, a layout-guided 3D Gaussian splatting pipeline designed for sparse panoramic images, which allows real-time panoramic rendering using our novel ${360^\circ}$ Gaussian splatting algorithm. 
    \item We derive a high-quality point cloud generation method for the initialization of 3D Gaussians from room layout priors to improve the performance of few-shot novel view synthesis.
    \item We introduce a layout-guided regularization on 3D Gaussians to reduce floaters caused by under-constrained regions.
\end{itemize}

\begin{figure*}[t]
  \centering
   \includegraphics[width=1.0\linewidth]{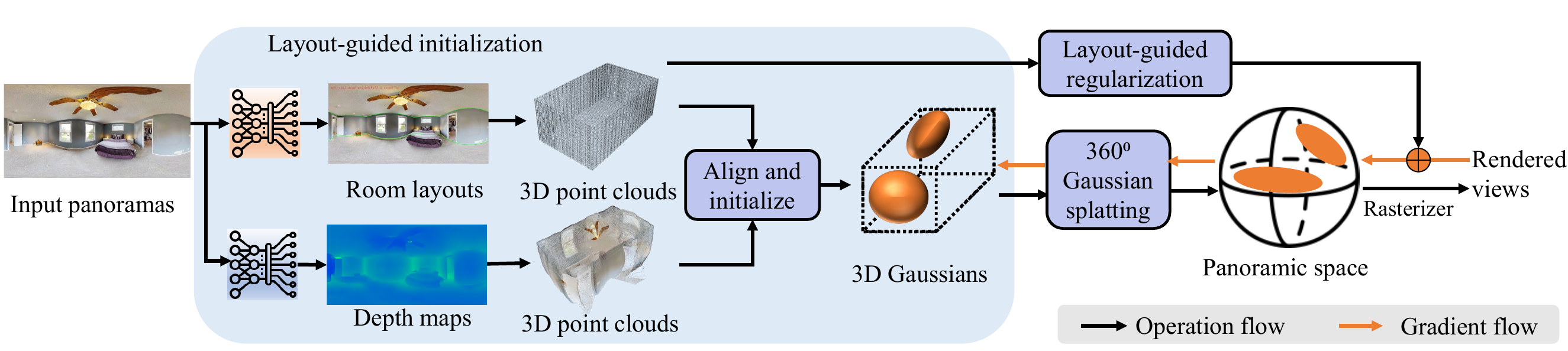}
   \caption{\textbf{Overview of 360-GS architecture}. Given a limited set of panoramas, we estimate the room layout and depth to guide the optimization of 3D Gaussian. These priors are transformed into 3D representations and jointly merged to a point cloud, which is used to initialize 3D Gaussians. We propose a 3D Gaussian splatting algorithm to project 3D Gaussians to panoramic space. Based on the projected Gaussians, we can render panoramas through a differentiable tile rasterizer. To reduce floaters in novel views, we regularize the optimization of 3D Gaussians by minimizing the cosine distance between the movement of position vectors and normals of layout point clouds.}
   \label{fig:pipeline}
\end{figure*}

\section{Related Work}
\subsection{Novel view synthesis}
Given a dense set of calibrated images, the task of novel view synthesis aims to generate photo-realistic images of a 3D scene from unseen viewpoints. To improve the quality of the reconstructed 3D scene and novel views, some studies utilize explicit representations such as layered representations~\cite{Shade1998LayeredDI, Shih20203DPU}, voxels~\cite{Sitzmann2018DeepVoxelsLP}, mesh~\cite{Jack2018LearningFD} and point clouds~\cite{Qi2016PointNetDL, Wiles2019SynSinEV}. Recently, there has been an increasing interest in the use of volumetric representations. Neural radiance fields (NeRF)~\cite{Mildenhall2020NeRFRS} employs implicit neural networks to represent scenes as continuous volumetric functions of density and color. Volumetric rendering is then employed to generate novel views. Mip-NeRF 360~\cite{Barron2021MipNeRF3U} extends NeRF to address aliasing and model unbounded scenes.
Despite the powerful neural implicit representation of NeRF, it demands significant time for training and rendering, posing a challenge for real-time applications. Recent works have strived to accelerate the rendering speed~\cite{Mller2022InstantNG, Reiser2021KiloNeRFSU, Sun2021DirectVG, Yu2021PlenoxelsRF}. Concurrently, another line of work employs point-based representation and rendering. 3D-GS~\cite{3DGS} models the scene with explicit 3D Gaussians and efficiently renders 2D images using the splatting technique, elevating the photo-realistic rendering quality to real-time levels.
While existing methods for novel view synthesis primarily focus on perspective images, recent research has been adapted for panoramic input. OmniNeRF~\cite{Gu2022OmniNeRFNR} extends the pinhole camera model of NeRF to a fish-eye projection model and uses spherical sampling to enhance the quality of rendering.
360Roam~\cite{Huang2022360RoamRI} is the first to construct an omnidirectional neural radiance field from a sequence of panoramic images. 360FusionNeRF~\cite{Kulkarni2022360FusionNeRFPN} introduces a semantic consistency loss to enforce 3D space consistency in panoramas. In contrast, recent works built on 3D Gaussian splatting barely consider panoramas as input. To our knowledge, we are the first to extend 3D-GS to panoramic view synthesis.

\subsection{Layout priors in panoramas}
Panoramic room layout estimation plays a crucial role in indoor scene comprehension and has been extensively studied~\cite{HorizonNet, Jiang2022LGTNetIP, Pintore2020AtlantaNetIT, Su2022GPRNetML, Shen_2023_CVPR,Shen2023DisentanglingOP}. Among them, HorizonNet~\cite{HorizonNet}  introduces a deep learning network and a post-processing technique that can recover complex room layouts, even with obscured corners from the model output. 
This estimated panoramic room layout has been widely explored in computer vision problems including indoor navigation~\cite{Mirowski2016LearningTN} and scene reconstruction~\cite{Izadinia2016IM2CAD}. 
In the task of novel view synthesis, Xu et al.~\shortcite{Xu2021LayoutGuidedNV} utilize the estimated room layout from the reference panorama and extract high-level features as guidance for target views, proving the efficacy of layout priors. However, the neural information of the room layout is underutilized in 3D Gaussians, as 3D Gaussians lack neural components. Unlike this method, we exploit room layout priors through explicit initialization and geometric constraints for 3D Gaussians.

\begin{figure}[t]
  \centering
   \includegraphics[width=1.0\linewidth]{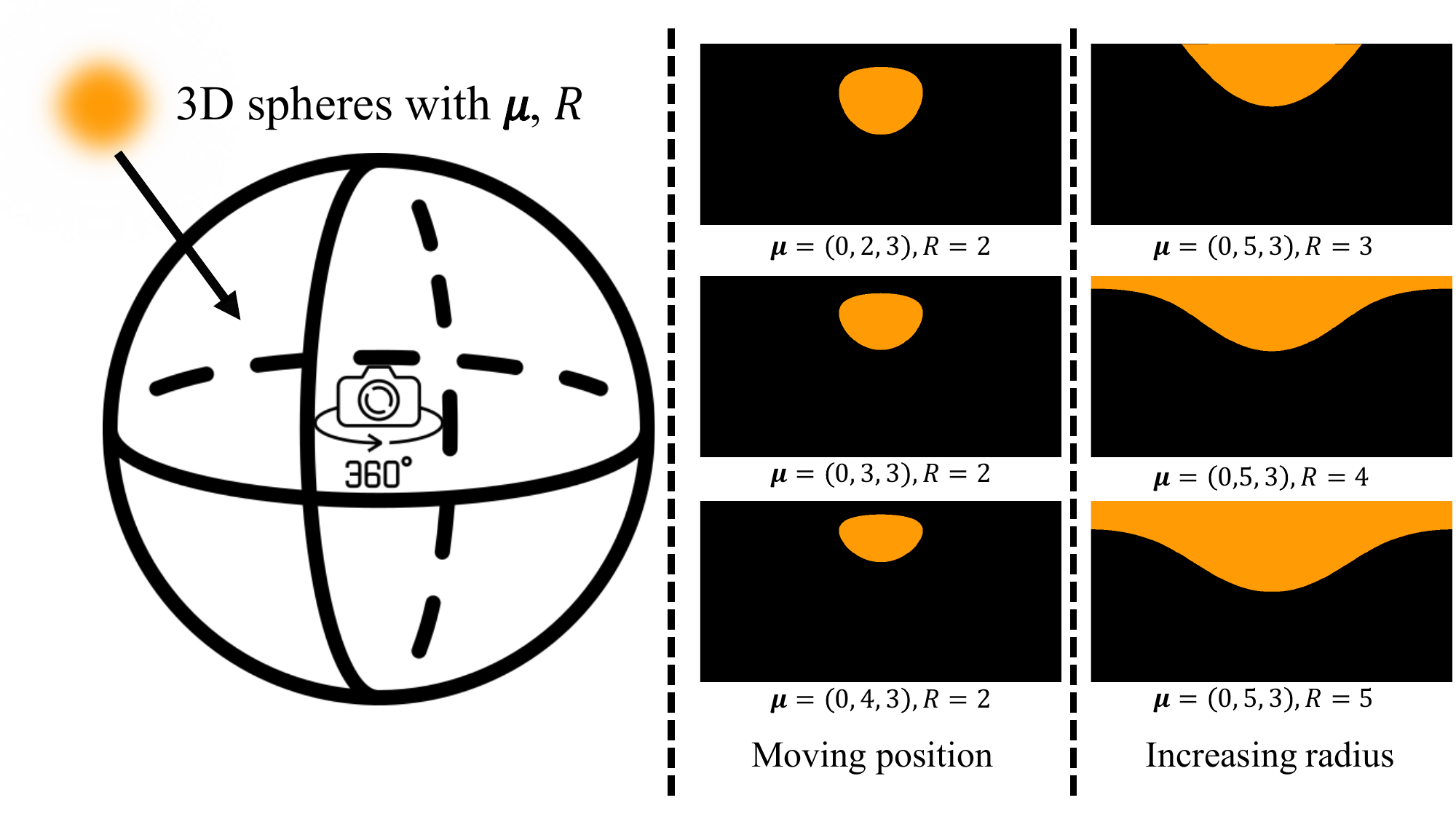}
   \caption{\textbf{Illustration of panoramic Gaussian splatting}. We show a toy case for splatting a 3D sphere (a special case of 3D Gaussian) onto panoramic images. The 3D sphere is defined by its position $\bm{\mu} \in \mathbb{R}^3$ and its radius $R \in \mathbb{R}$. In the middle column, we visualize the projection of 3D spheres under different positions. In the right column, 3D spheres with varying radii are splatted onto the panoramas. These projections are not elliptical and cannot be accurately modeled with 2D Gaussians, as attempted by 3D-GS.}
   \label{fig:panoramic_projection}
\end{figure}
\begin{figure}[htp]
  \centering
   \includegraphics[width=1.0\linewidth]{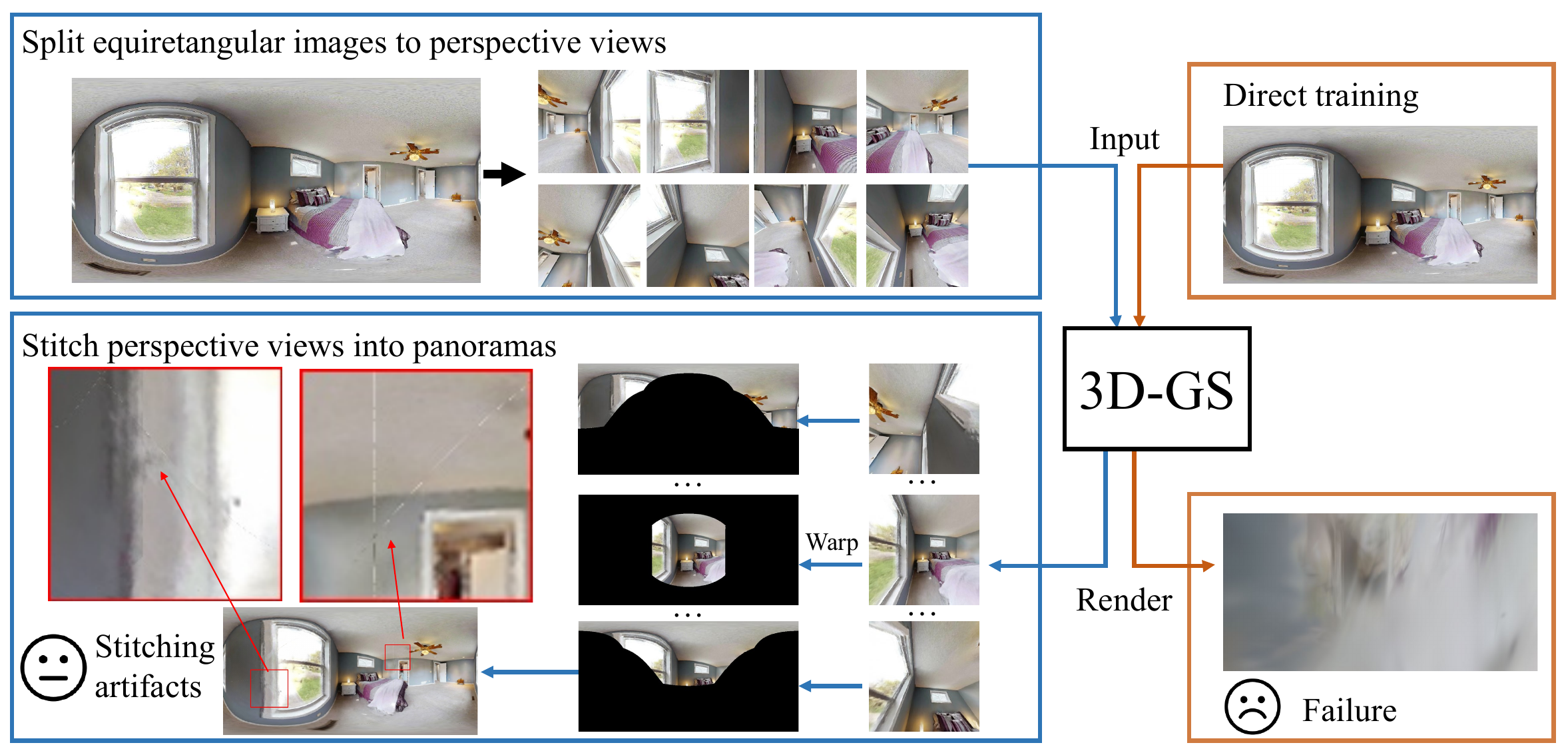}
   \caption{\textbf{Two naive pipelines for applying 3D-GS to panoramic inputs}. Left: we split panoramas into perspective images with poses and feed them to 3D-GS. To render panoramas, we render perspective images centered at the camera of panoramas. Subsequently, these images are transformed and stitched into equirectangular projection. These operations introduce stitching artifacts. Right: we fail to directly train 3D-GS with panoramic inputs.}
   \label{fig:eq2fov}
\end{figure}
\section{Method}
We present 360-GS, a pipeline designed to optimize 3D Gaussians and facilitate panoramic rendering. Fig.~\ref{fig:pipeline} shows an overview of our 360-GS. We identify challenges in adapting panoramas to 3D-GS (Sec.~\ref{sec:challenge}) and propose ${360^\circ}$ Gaussian splatting as a solution. 360-GS further designs a layout-guided initialization and regularization to fully exploit room layout priors within panoramic input.
\subsection{Preliminary and challenge} \label{sec:challenge}
3D-GS~\cite{3DGS} explicitly represents a 3D scene with a collection of 3D Gaussians in world space. Each Gaussian is defined by a position vector $\bm{\mu} \in \mathbb{R}^3$ and a covariance matrix $\Sigma \in \mathbb{R}^{3\times 3}$. The 3D Gaussian distribution can be represented as follows:
\begin{equation}
G(\bm{x})=e^{-\frac{1}{2}(\bm{x}-\bm{\mu})^T \Sigma^{-1}(\bm{x}-\bm{\mu})}
\end{equation}
where $\Sigma$ can be described with a scaling matrix $\bm{S}$ and a rotation matrix $\bm{R}$ as follows:
\begin{equation}
\Sigma= \bm{R}\bm{S}\bm{S}^{T}\bm{R}^{T}.
\end{equation}

For differentiable optimization, 3D-GS renders 2D images by projecting 3D Gaussians to 2D image planes. Given points $\bm{x}=(x_0,x_1,x_2)$ in world coordinates, we first transform them to camera coordinates $\bm{t}=(t_0,t_1,t_2)$ using an affine mapping $\bm{t}=V(\bm{x})=\bm{W}\bm{x}+\bm{d}$, known as the viewing transformation. Subsequently, the camera coordinates are converted to ray coordinates through the mapping $\bm{x} = p(\bm{t})$. 

Taking perspective images as inputs, these mappings are in fact not affine. To solve this problem, Zwicker et al.~\shortcite{EWA} introduce the local affine approximation of the projective transformation with the Jacobian matrix $\bm{J}$. As a result, the new covariance matrix $\Sigma^{^{\prime}}$ of projected 2D Gaussians in camera coordinates are formulated as:
\begin{equation}
\Sigma^{^{\prime}}=\bm{J}\bm{W}\Sigma \bm{W}^{T}\bm{J}^{T}.
\end{equation}

3D-GS~\cite{3DGS} skips the third row and column of $\Sigma^{\prime}$, resulting in a $2\times 2$ variance matrix with the same structure and properties. Therefore, the projection of 3D Gaussians is represented with 2D Gaussians.

Since the local affine approximation relies on projective transformation, it is not suitable for mapping 3D Gaussians to 2D Gaussians on panoramic images. A panoramic image covers the whole ${360^\circ}$ horizontally and the whole ${180^\circ}$ vertically. Consequently, the top and bottom of the image appear severely distorted. As illustrated in Fig.~\ref{fig:panoramic_projection}, the panoramic projection assumes distinct shapes that can not be modeled with Gaussians under varying configurations. Employing a 2D Gaussian for fitting such a projection would lead to significant errors.

An alternative approach for the application of 3D-GS to panoramic inputs involves transforming the panoramas into perspective images before optimizing 3D Gaussians. An overview of this method is illustrated in Fig.~\ref{fig:eq2fov}. Concretely, we split equirectangular images into $N$ perspective views, each associated with a distinct pose. Then these perspective images can be utilized to optimize 3D Gaussians following 3D-GS. However, this straightforward solution presents two main drawbacks:  (1) The complete pipeline is intricate, and the direct acquisition of panoramas is unfeasible. (2) To get a complete panorama, more than six perspective images are supposed to be rendered and concatenated jointly. Unfortunately, this concatenation introduces inevitable stitching artifacts in overlapping regions of the reconstructed panoramas shown in Fig.~\ref{fig:eq2fov}.
\begin{figure}[t]
  \centering
   \includegraphics[width=1.0\linewidth]{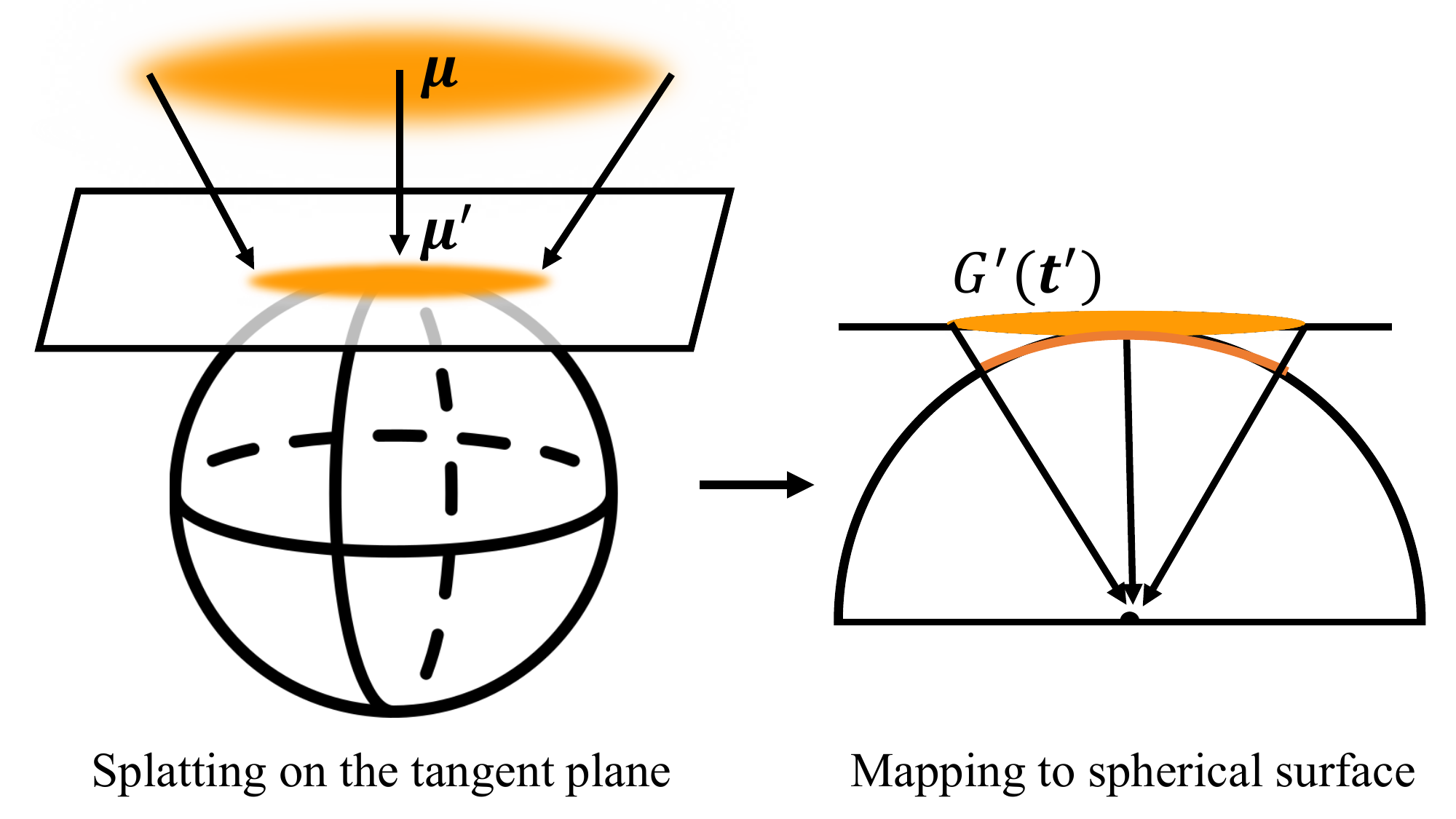}
   \caption{\textbf{${360^\circ}$ Gaussian splatting algorithm.} ${360^\circ}$ Gaussian splatting first splats 3D Gaussians on the tangent plane that passes through the projection point $\bm {\mu}^{\prime}$, yielding the distribution $G^{\prime}(\bm{t}^{\prime})$. Then we map the projection to the spherical surface of the unit sphere. }
   \label{fig:splat}
\end{figure}
\subsection{${360^\circ}$ Gaussian splatting}\label{sec:360splatting}
Our goal is to optimize 3D Gaussian representations from a set of panoramas and enable direct panorama rendering. Considering the challenges of directly representing spherical projection, we introduce a novel splatting technique that decomposes the splatting on the spherical surface into two sequential steps: splatting on the tangent plane of the unit sphere and mapping to the spherical surface. This allows us to project 3D Gaussians to 2D Gaussians for rendering. An overview of our ${360^\circ}$ Gaussian splatting is illustrated in Fig.~\ref{fig:splat}. 

Given a 3D elliptical Gaussian centered at $\bm{\mu}$ with a covariance matrix $\Sigma$, we first convert it to the camera coordinates with the affine viewing transformation $V(\bm{x})$. The viewing transformation is then followed by a projective transformation $\bm{t^{\prime}}= \varphi(\bm{t}, \bm{\mu^{\prime}})$ that projects camera coordinates to the tangent plane of the unit sphere. This tangent plane passes through the projection point $\bm{\mu^{\prime}}$ and is tangential to the unit sphere centered at the origin of the camera coordinates. The transformation is formulated as:
\begin{equation}
\bm{\mu^{\prime}}(\bm{x} -\bm{\mu^{\prime}}) = 0
\end{equation}
where $\bm{\mu^{\prime}}$ is the projection of $V(\bm{\mu})$ onto the unit sphere. Thus the projection is given by:
\begin{equation}
(t^{^{\prime}}_0,  t^{^{\prime}}_1, t^{\prime}_2)^T
=\varphi(\bm{t},\bm{\mu^{\prime}}) = \bm{t} \frac{ (\bm{\mu ^{\prime}})^T \bm{\mu ^{\prime}}}{(\bm{\mu^{\prime}})^T \bm{t}}.
\end{equation}

Following Zwicker et al.~\shortcite{EWA}, we define the local affine approximation $\varphi_{k}(\bm{t},\bm{\mu^{\prime}})$ by the first two terms of the Taylor expansion of $\varphi$ at the point $\bm{t}_k$:
\begin{equation}
\varphi_k(\bm{t},\bm{\mu^{\prime}}) = \varphi_k(\bm{t_k},\bm{\mu^{\prime}}) +\bm{J}_k \cdot (\bm{t} -\bm{t}_k)
\end{equation}
where $\bm{t}_k=(t_0, t_1, t_2)^T=V(\bm{\mu})$ is the center of the 3D Gaussian in camera coordinates. The Jacobian $\bm{J}_k$ is given by the partial derivatives  of $\varphi$ at the point $\bm{t}_k$:

\begin{equation}
\bm{J}_k = \frac{-1}{\left(\mu^{\prime}_{0} t_0 + \mu^{\prime}_{1} t_1 + \mu^{\prime}_{2} t_2\right)^{2}}\left[\begin{matrix}
{\mu^{\prime}_{1} t_1 + \mu^{\prime}_{2} t_2}
& \mu^{\prime}_{1} t_0
& \mu^{\prime}_{2} t_0\\
\mu^{\prime}_{0} t_1 
&\mu^{\prime}_{0} t_0 + \mu^{\prime}_{2} t_2
& \mu^{\prime}_{2} t_1\\
\mu^{\prime}_{0} t_2 
& \mu^{\prime}_{1} t_2 
& \mu^{\prime}_{0} t_0 + \mu^{\prime}_{1} t_1
\end{matrix}
\right].
\end{equation}
We splat 3D Gaussians onto the tangent plane by concatenating $\bm{t}= V(\bm{x})$ and $\bm{t^{\prime}}= \varphi(\bm{t}, \bm{\mu^{\prime}})$, yielding the distribution as follows:
\begin{equation}
G^{\prime}(\bm{t}^{\prime})=e^{-\frac{1}{2}(\bm{t}^{\prime}-\bm{\mu}^{\prime})^T (\bm{J}_k\bm{W}\Sigma\bm{W}^T{\bm{J}_k}^T)^{-1}(\bm{t}^{\prime}-\bm{\mu}^{\prime})}.
\end{equation}

For panorama rendering, we map the tangent plane in camera coordinates to the spherical surface in spherical polar coordinates. The mapping $(\theta, \phi) = P_{360}(\bm{t}^{\prime})$ is therefore given by: 
\begin{equation}
\left (
\begin{matrix}
   \theta \\
   \phi 
\end{matrix}
\right )= P_{360}(\bm{t}^{\prime})=
\left(
\begin{matrix}
    \text{atan2} (-t_1, \sqrt{t_0^2+t_2^2})\\
    \text{atan2} (t_0, t_2) 
\end{matrix}
\right )
\end{equation}
where $\bm{t}^{\prime}=(t^{^{\prime}}_0,  t^{^{\prime}}_1, t^{\prime}_2)^T$ is a point on the tangent plane and $(\theta, \phi)$ denote the latitude and longitude of the spherical surface. The spherical surface and the panoramic pixel grid coordinates are related by:
\begin{equation}
\left\{
\begin{aligned}
r &= -{\theta *H }/{\pi}+{H}/{2}\\
c &= {\phi *W }/{2\pi}+{W}/{2}
\end{aligned}
\right.
\end{equation}
where $(r, c)$ is the row and column of the panorama with the resolution of $H\times W$.

Through projection and one-to-one mapping, 3D Gaussians are splattered onto the panoramas. Since the mapping process is efficient, our approach maintains real-time performance. Subsequently, the pixel color of the panorama is derived through alpha-blending these layered splattered Gaussians from front to back, following 3D-GS. Consequently, we are enabled to render panoramas directly.

\subsection{Layout prior for panoramas}
In the context of sparse panoramas lacking 3D information, 3D-GS struggles to identify cross-view 3D correspondences and construct the geometry of scenes, leading to a significant degradation in the quality of novel view synthesis.
In this paper, we exploit the room layout, a form of 3D structural information within panoramas, to alleviate these issues. Incorporating the room layout with 3D Gaussians has three advantages. First, it contains whole-room contextual information and 3D priors, which are consistent across diverse views. Second, different from depth maps and point-cloud representations, the room layout describes the scene with a smooth surface structure, yielding seamless planes including walls and floors~\cite{Jiang2022LGTNetIP}. Third, room layouts are easily accessible and robust to the scale of scenes. Recent advancements have significantly propelled the field of layout estimation, attaining frame rates exceeding 20 frames per second (FPS)~\cite{HorizonNet}.

Under the assumption that room layouts conform to the Atlanta World assumption~\cite{Pintore2020AtlantaNetIT}, room layouts are composed of vertical walls, horizontal floor and ceiling. 
As illustrated in Fig.~\ref{fig:pipeline}, we depict room layouts using floor-wall boundaries $\bm{B}_{f} \in \mathbb{R}^{1\times W}$ and ceiling-wall boundaries $\bm{B}_{c} \in \mathbb{R}^{1\times W}$, where $\bm{B}_{f} $ and $\bm{B}_{c}$ represent the latitudes of ceiling and floor boundaries of each image column, respectively.
With the known camera height, we transform the 2D boundaries from images to 3D positions. Subsequently, we recover floor, ceiling, and wall planes following Pintore et al.~\shortcite{HorizonNet}, thus creating a 3D bounding box as the 3D room layout of the scene.

\subsection{Layout-guided initialization}
Previous studies~\cite{3DGS, Chen2023Textto3DUG} have demonstrated the importance of a reasonable geometric initialization in training 3D Gaussians. 3D-GS advocates for starting with an initial set of sparse points derived from Structure-from-Motion (SfM)~\cite{schoenberger2016sfm, schoenberger2016mvs}. However, Sfm fails with sparse-view inputs, and so cannot reliably provide point cloud initializations~\cite{Sinha2022SparsePoseSC}.

Given that room layouts reveal the global geometric structure of the scene, we integrate the layout point cloud into the initialization. Specifically, we estimate a floor-wall boundary and ceiling-wall boundary as the layout for each panorama using off-the-shelf network~\cite{HorizonNet}. To obtain the global layout of the scene, we merge the boundaries of all panoramas using a 2D union operation. Subsequently, we construct the corresponding 3D bounding box from the global layout and then convert it into a point cloud through uniform sampling. 
To augment information for objects not included in the layout, we also estimate depth for panoramas and convert depth maps into point clouds~\cite{Pintore2021SliceNetDD}. We merge these depth point clouds and then downsample it to reduce the number of points while maintaining the structure of objects. The merged depth point cloud is aligned to the layout point cloud with a global scale factor. Finally, we combine the layout and depth point clouds to initialize 3D Gaussians. 

\begin{figure}[t]
  \centering
   \includegraphics[width=1.0\linewidth]{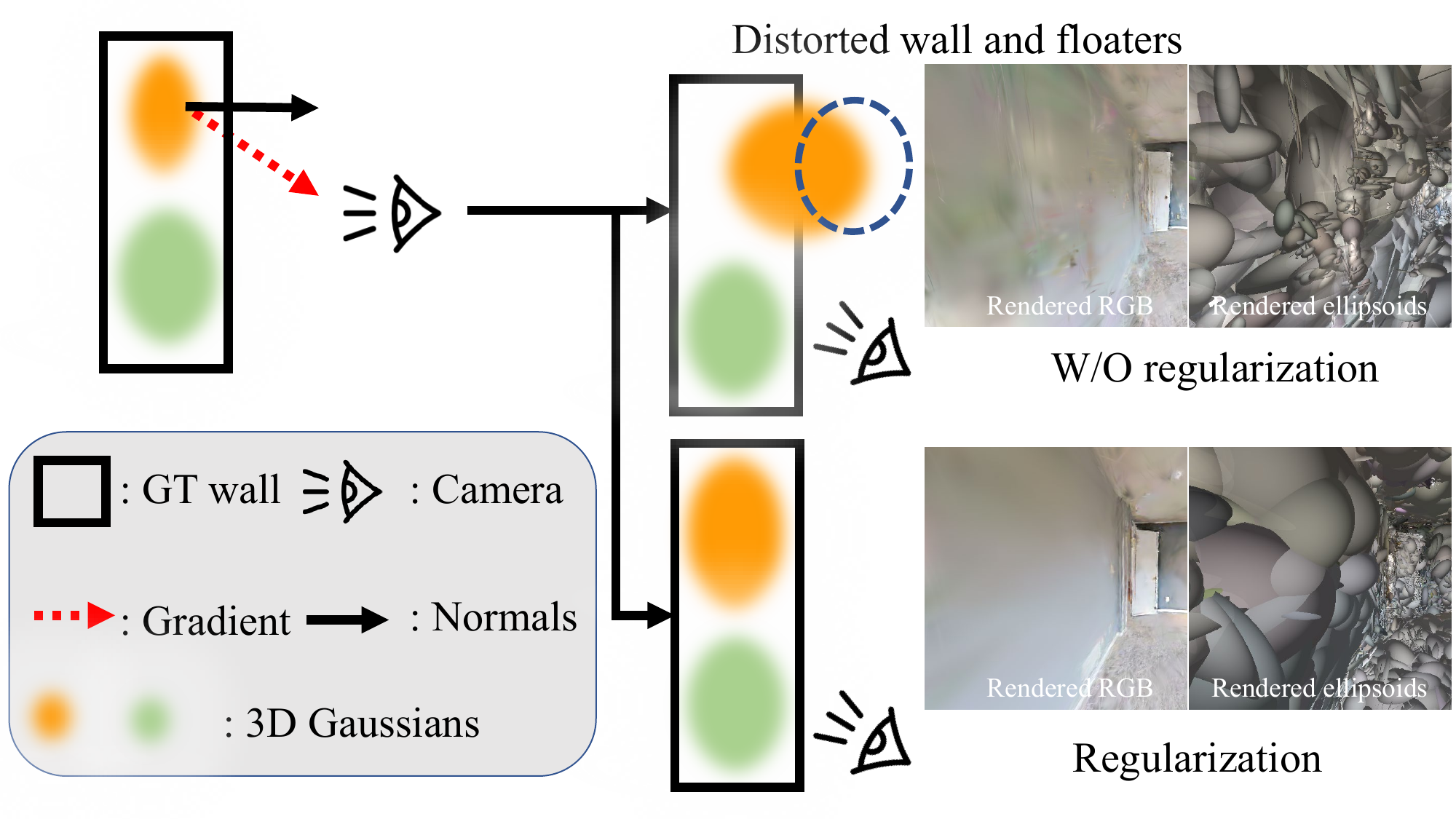}
   \caption{\textbf{Impact of layout-guided regularization.} We present a 2D toy case for optimizing 3D Gaussians, marked with an orange color. Without our regularization, 3D Gaussians gravitate towards the gradient direction, disrupting the inherent layout structure. This results in some Gaussians appearing outside the walls, causing distorted planes and ``floaters" in novel views. Our layout-guided regularization effectively preserves the overall structure during optimization.}
   \label{fig:regularization}
\end{figure}
\subsection{Layout-guided regularization}
Although 3D Gaussians are initially set with a layout-guided point cloud, the room layout priors within panoramas suffer from catastrophic forgetting. As illustrated in Fig~\ref{fig:regularization}, the parameters of 3D Gaussians, such as position vectors $\bm{\mu}$, are optimized in the direction of the gradient. Consequently, 3D-GS struggles to preserve the geometric structure initialized with layout priors, leading to uneven surfaces and the emergence of ``floaters" in novel views.

To address this issue, we introduce a layout-guided regularization to enforce 3D Gaussians to maintain the consistency of the room layout. Specifically, when we initialize 3D Gaussians with the layout point cloud, we additionally record their corresponding initial positions $\bm{u_0}$ and normals $\bm{n}$ of 3D layout points. We regularize the optimization of 3D Gaussians by minimizing the cosine distance between the movement of the position and the normals. We then aggregate these cosine distances for all 3D Gaussians that are initialized with layout point clouds. The final layout-guided loss is formulated as:
\begin{equation}
\mathcal{L}_{\text{layout}} =\sum \frac{\bm{n} \cdot (\bm{\mu}-\bm{u_0})}{\Vert \bm{n}\Vert 
 \times \Vert \bm{\mu}-\bm{u_0}\Vert}.
\end{equation}

\begin{table}[t]
\begin{center}
\caption{Quantitative evaluation of our method against \colorbox{LightGreen}{NeRF-based} methods and \colorbox{LightBlue}{3D-GS} with different initialization strategies. The \textbf{best} and \underline{second-best} scores are highlighted, respectively. 3D-GS$^{*}$ refers to the random initialization method and 3D-GS is initialized with Sfm point clouds. Note that Sfm point clouds are not available in the sparse 4-view inputs, resulting in some blank sections. }
\label{tab:results}
\begin{tabular}{c|c|c|c|c|c|c}
\multicolumn{2}{c|}{Metrics} & \cellcolor{LightGreen}M-360 & \cellcolor{LightGreen}INGP & \cellcolor{LightBlue}3D-GS$^{*}$ & \cellcolor{LightBlue}3D-GS &\cellcolor{LightRed}Ours\\ \toprule
\multicolumn{2}{c|}{FPS} & 0.07& 3.08&\multicolumn{2}{c|}{60} & 60\\ \hline
\multirow{3}{*}{\rotatebox{90}{4-view}}&PSNR$\uparrow$ &\textbf{19.15} &15.49 &13.92 &-&\underline{18.96}\\
&SSIM$\uparrow$ &\textbf{0.633} &0.432 & 0.438&- &\underline{0.600}\\
&LPIPS$\downarrow$ &\underline{0.374} & 0.586& 0.547&-&\textbf{0.344}\\ \midrule
\multirow{3}{*}{\rotatebox{90}{32-view}}&PSNR$\uparrow$ &26.72 &\textbf{28.23} &21.65 &26.74 &\underline{28.22}\\
&SSIM$\uparrow$ &0.835 &\underline{0.860} & 0.704&0.837 &\textbf{0.871}\\
&LPIPS$\downarrow$ & 0.186& \textbf{0.099}&0.334 &0.168 &\underline{0.107}\\
 \bottomrule
\end{tabular}
\end{center}
\end{table}

\section{Experiments}

\subsection{Implementation details}
Our 360-GS is implemented based on the Pytorch framework in 3D-GS~\cite{3DGS}. To obtain priors for our layout-guided initialization, we utilize the pretrained HorizonNet~\cite{HorizonNet} for layout estimation and SliceNet~\cite{Pintore2021SliceNetDD} for monocular panoramic depth estimation.
Our final loss function for optimization is defined as:
\begin{equation}
\mathcal{L}=\lambda_1 \Vert \bm{C}-\bm{\hat{C}} \Vert_1+\lambda_2 \mathcal{L}_{\text{D-SSIM}}+\lambda_3 \mathcal{L}_{\text{layout}}
\end{equation}
where $\mathcal{L}_{\text{D-SSIM}}$ is the D-SSIM term between rendered panoramas $\bm{C}$ and ground truth panoramas $\bm{\hat{C}}$. $\mathcal{L}_{\text{layout}}$ stands for the layout-guided regularization terms. 
For 4-view inputs, we set $\lambda_1$, $\lambda_2$, and $\lambda_3$ as 0.8, 0.2, and 0.1 respectively. For 32-view inputs. $\lambda_3$ is set to 0.01 to better fit the sufficient inputs.

\subsection{Experimental setting}
\textit{Dataset.}
For both quantitative and qualitative evaluations, we gathered a total of 10 real-world scenes from the publicly available Matterport3D dataset~\cite{Chang2017Matterport3DLF}. Each scene, characterized by varied styles and furniture configurations, contains over 40 panoramas, each with a resolution of $512\times 1024$ pixels. From these panoramas, we uniformly selected 4 and 32 panoramas as the training views for each scene. The remaining panoramas constitute the test set.

\textit{Baseline and metrics.}
We compare 360-GS with 3D-GS~\cite{3DGS} and two state-of-the-art NeRF-based approaches: MipNeRF-360 (M-360)~\cite{Barron2021MipNeRF3U} and Instant NGP (INGP)~\cite{Mller2022InstantNG}. 
Given that 3D-GS only processes perspective images, we split each training panorama into eight perspective images, each with a resolution of $512\times 512$ pixels. As suggested by Tancik et al.~\shortcite{nerfstudio}, we assume a camera field of view of 120 degrees and capture perspective images horizontally at the elevation angles of $[-45^\circ,0^\circ,45^\circ]$ with this camera. For testing, 3D-GS generated eight images from test viewpoints and combined them to form panoramas. As MipNeRF-360 currently leads in NeRF rendering quality for perspective images, we trained it with perspective images and evaluated it on panoramas. We adapted INGP, a recent real-time rendering NeRF, for panoramic input following Huang et al.~\cite{Huang2022360RoamRI}, to reduce the time for rendering panoramas. We train MipNeRF-360 for 250k iterations using the official code, which takes approximately 12 hours. INGP runs for 30 epoches, taking about 12 minutes per scene. Both 3D-GS and 360-GS are trained for 7k iterations with default parameters in the official code of 3D-GS. All our experiments are conducted on a single GPU Nvidia RTX 3090. 
For evaluation, we report the average PSNR, SSIM, and LPIPS scores for all the methods under different numbers of training views. In addition, we report the FPS for rendering a $512\times 1024$ image.

\begin{figure}[t]
  \centering
   \includegraphics[width=1.0\linewidth]{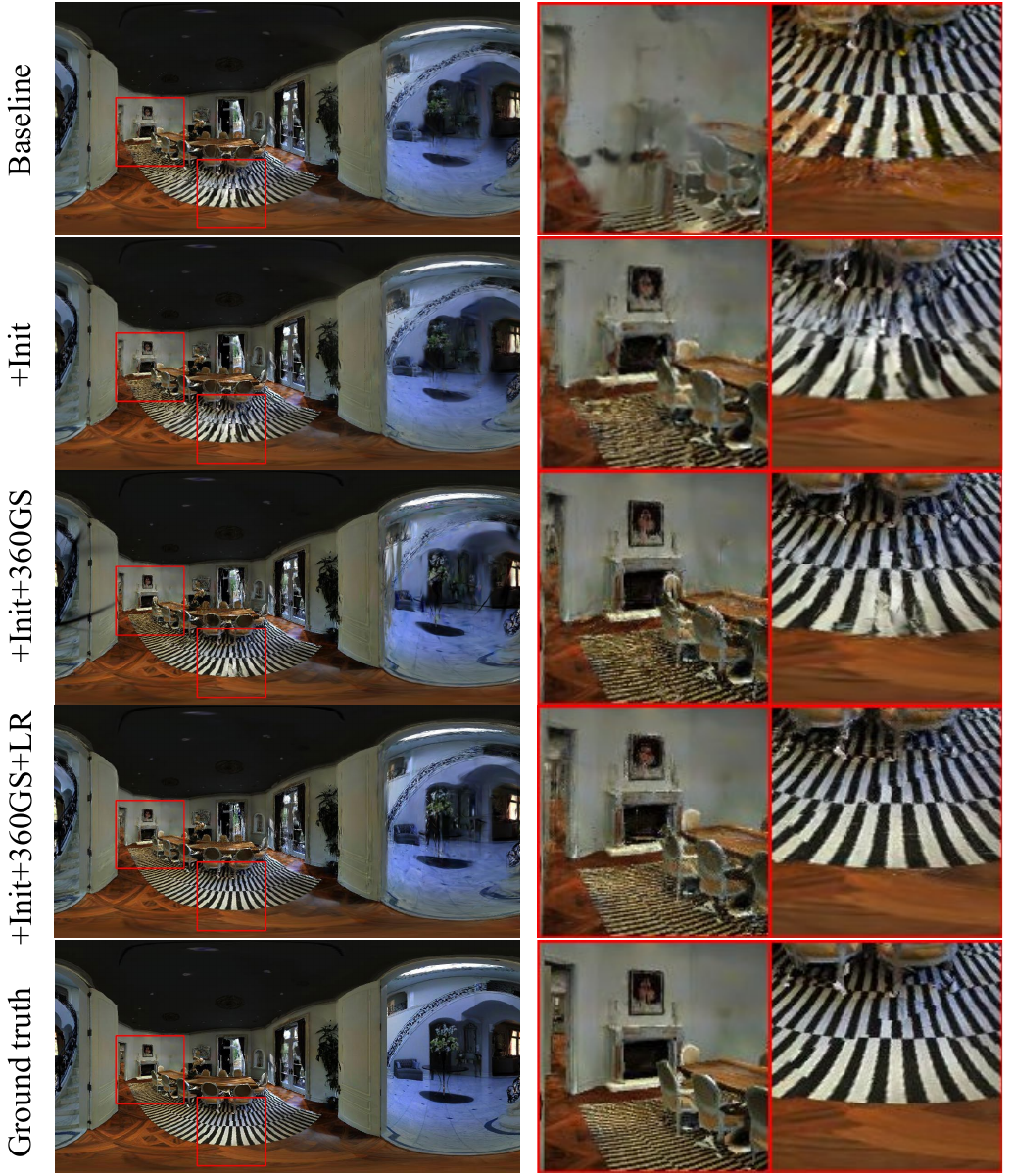}
   \caption{\textbf{Visualization of ablation study.} }
   \label{fig:ablation}
\end{figure}

\subsection{Results}

\textit{Quantitative comparisons.}
Tab.~\ref{tab:results} reports the quantitative results of SOTA methods and our 360-GS. Our method outperforms 3D-GS in terms of all metrics and input settings. In a 4-view setting, our method surpasses 3D-GS$^{*}$ with a remarkable 5.04 PSNR improvement. Despite the substantial performance improvement of 3D-GS with Sfm point cloud initialization, it still falls short when compared to our method due to stitching artifacts.
In the 32-view evaluation, INGP excels with the highest PSNR and LPIPS, while our proposed method leads in SSIM. This showcases the competitiveness of our method with INGP, especially considering that INGP's performance dramatically degrades in the 4-view setting. In the 4-view evaluation, our method is comparable with MipNeRF-360 and outperforms other methods in terms of LPIPS. However, the training and rendering time for MipNeRF-360 is considerably longer than ours. 
The quantitative comparison demonstrates that our method achieves state-of-the-art performance while ensuring fast rendering and robustness to the limited views.

\begin{figure*}[t]
  \begin{center}
  \includegraphics[width=\linewidth, trim={0px, 0px, 0px, 0px}, clip]{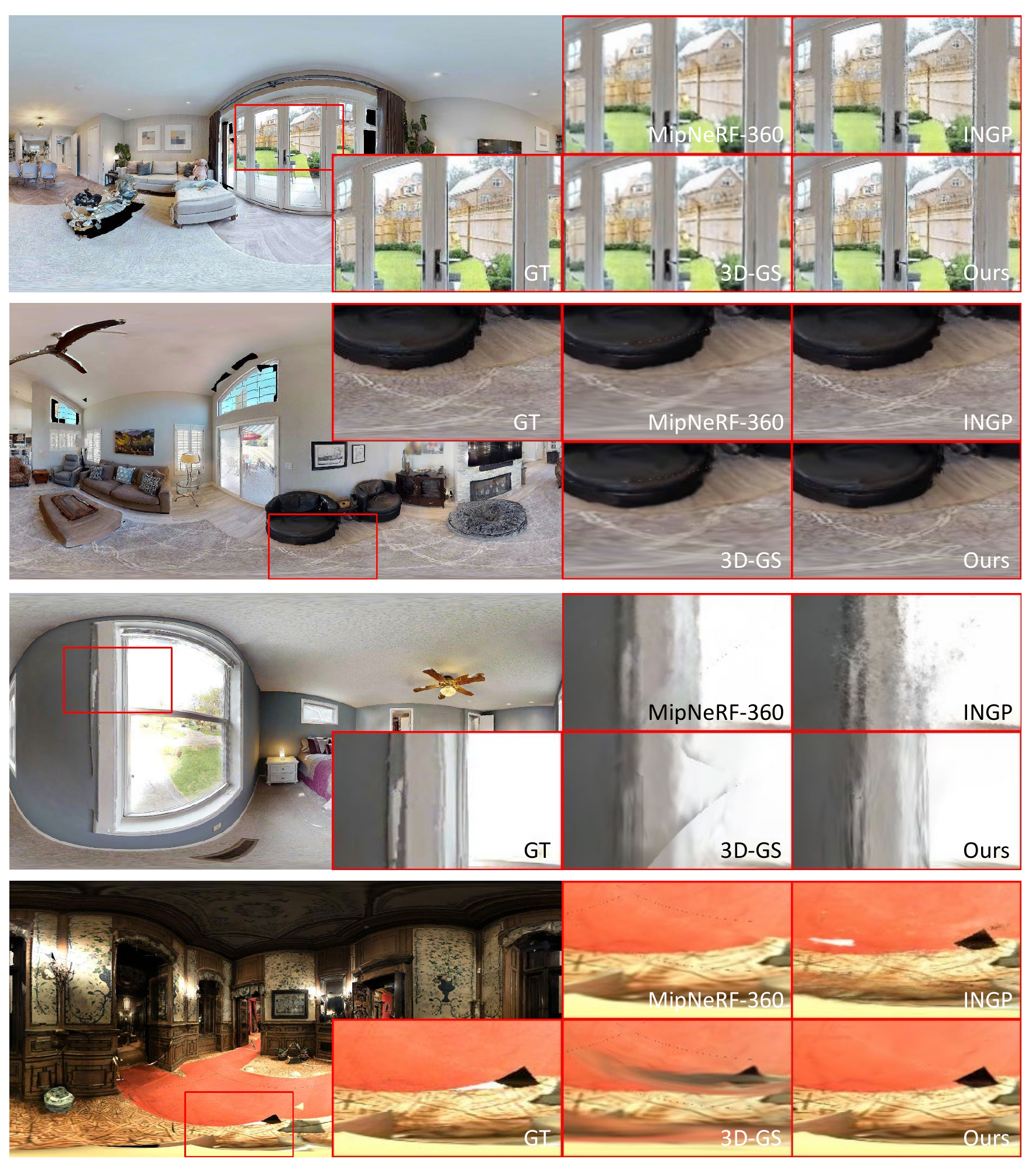}
  \end{center}
  \caption{Qualitative comparison of our methods and some SOTA methods with 32-view inputs. 3D-GS needs to stitch perspective images into a panorama, resulting in stitching artifacts shown in the third row. Our method circumvents these artifacts and faithfully produces texture on the planes such as walls and floors. This is attributed to our exploitation of room layout priors and effective panoramic Gaussian splatting design.}
  \label{fig:32views}
\end{figure*}

\begin{figure*}[t]
  \begin{center}
  \includegraphics[width=\linewidth, trim={0px, 0px, 0px, 0px}, clip]{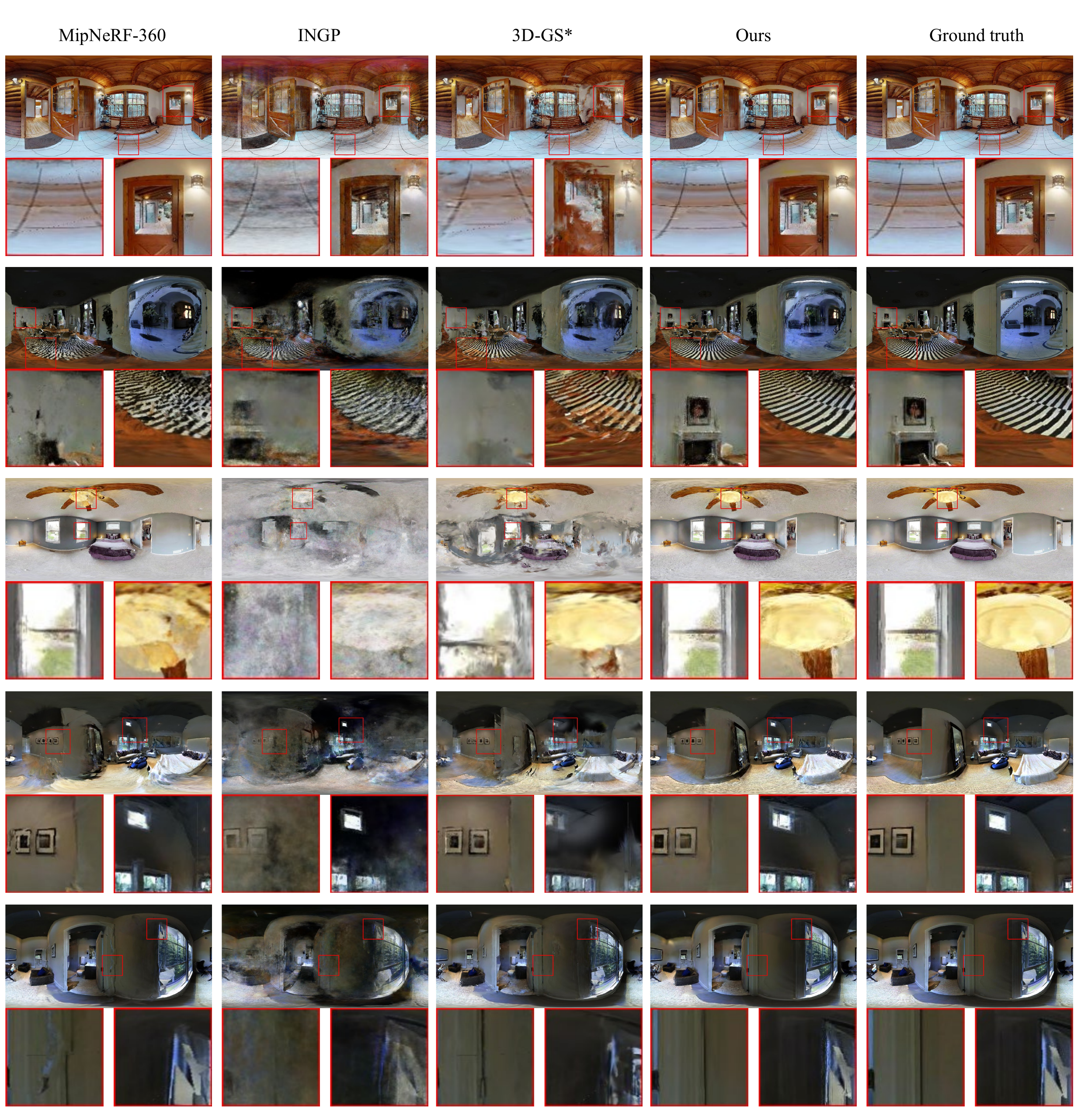}
  \end{center}
  \caption{Qualitative comparison of our methods and some SOTA methods with 4-view inputs. Due to the unavailability of the SfM point cloud in sparse views, we only present results of 3D-GS$^{*}$ initialized with a random point cloud.  The random initialization introduces noises and blurry artifacts in such under-constrained cases. INGP suffers from inadequate views, leading to over-smoothed outcomes and a tendency to overfit on training views. In contrast, our method produces visually appealing renderings comparable to those of MipNeRF-360. Moreover, our method holds the advantage of faster rendering while preserving better details.}
  \label{fig:4views}
\end{figure*}

\textit{Qualitative comparisons.} 
We present a qualitative comparison of the rendering results across all methods in Fig.~\ref{fig:32views} and Fig.~\ref{fig:4views}. Our method exhibits superior visual quality with 4-view inputs. Our results are comparable to those from MipNeRF-360 but at a lower computational cost. 360-GS effectively reconstructs the overall scene structure under the guidance of room layout priors, delivering visually pleasing results at first glance. Given a sufficient input of 32 panoramas, all methods yield satisfactory results. However, our method excels in recovering intricate patterns on planes, such as the floors depicted in the second and fourth rows of Fig.~\ref{fig:32views}.

\begin{table}[t]
\begin{center}
\caption{\textbf{Ablation studies.} We start with the baseline 3D-GS (first row), then incorporate our layout-guided initialization denoted as ``Init", resulting in a significant enhancement (second row).  The application of $360^{\circ}$ Gaussian splatting (360GS) further enhances the quantitative results in the third row. 360-GS is refined towards an improved solution with layout-guided regularization (LR) in the fourth row.}
\label{tab:ab_tab}
\begin{tabular}{c|c|c|c|c|c}

{Init} &{360GS}& {LR}& {PSNR$\uparrow$}& {SSIM$\uparrow$}   & {LPIPS$\downarrow$}\\
\toprule
\textcolor{red}{\XSolidBrush} & \textcolor{red}{\XSolidBrush} & \textcolor{red}{\XSolidBrush}& 13.64 & 0.458  & 0.459 \\
\textcolor{green}{\Checkmark} & \textcolor{red}{\XSolidBrush} & \textcolor{red}{\XSolidBrush}&  15.98 & 0.571  & 0.385 \\
\textcolor{green}{\Checkmark} & \textcolor{green}{\Checkmark} & \textcolor{red}{\XSolidBrush}& 16.66&0.588 & 0.334 \\
\textcolor{green}{\Checkmark} & \textcolor{green}{\Checkmark} & \textcolor{green}{\Checkmark}& 17.72&  0.622&0.318  \\
 \bottomrule
\end{tabular}
\end{center}
\end{table}

\subsection{Ablation study}
In Tab.~\ref{tab:ab_tab}, we validate the effectiveness of our design choice on a scene from the Matterport3D dataset under the 4-view inputs. We visualize the ablation results in Fig.~\ref{fig:ablation}.

\textit{Layout-guided initialization.} 
The baseline, 3D-GS, initialized with random point clouds, presents a substantial challenge for 3D Gaussians in learning the scene’s geometry. This results in novel views exhibiting noise and blurry artifacts due to the under-constrained random 3D Gaussians shown in the first row of Fig.~\ref{fig:ablation}. In contrast, the integration of layout-guided initialization into the baseline offers a plausible geometry for 3D Gaussians, leading to a significant PSNR enhancement of 2.34. This also aids in visually reconstructing the scene’s overall structure and wall-adjacent details like the fireplace, as depicted in the second row of Fig.~\ref{fig:ablation}.

\textit{$360^{\circ}$ Gaussian splatting.}
The performance of 3D-GS on panoramas is limited by stitching artifacts that arise during the concatenation process. Our $360^{\circ}$ Gaussian splatting fundamentally addresses this issue, leading to enhancements across all quantitative metrics as demonstrated in Tab.~\ref{tab:ab_tab}. As shown in the third row of Fig.~\ref{fig:ablation}, $360^{\circ}$ Gaussian splatting not only eliminates the stitching artifacts but also enriches the structural and visual details on the checkerboard-patterned carpet.

\textit{Layout-guided regularization.}
While room layout priors offer a plausible initial state for 3D Gaussians, there might be inconsistencies between optimized 3D Gaussians and the layout, leading to some artifacts on the carpet. In the fourth row of Fig.~\ref{fig:ablation}, we observe that our layout-guided regularization efficiently eliminates these artifacts, ensuring planes that are more consistent with geometric coherence. The effectiveness of the regularization is further demonstrated by a notable PSNR improvement of 1.06.

\begin{figure}[t]
  \centering
   \includegraphics[width=1.0\linewidth]{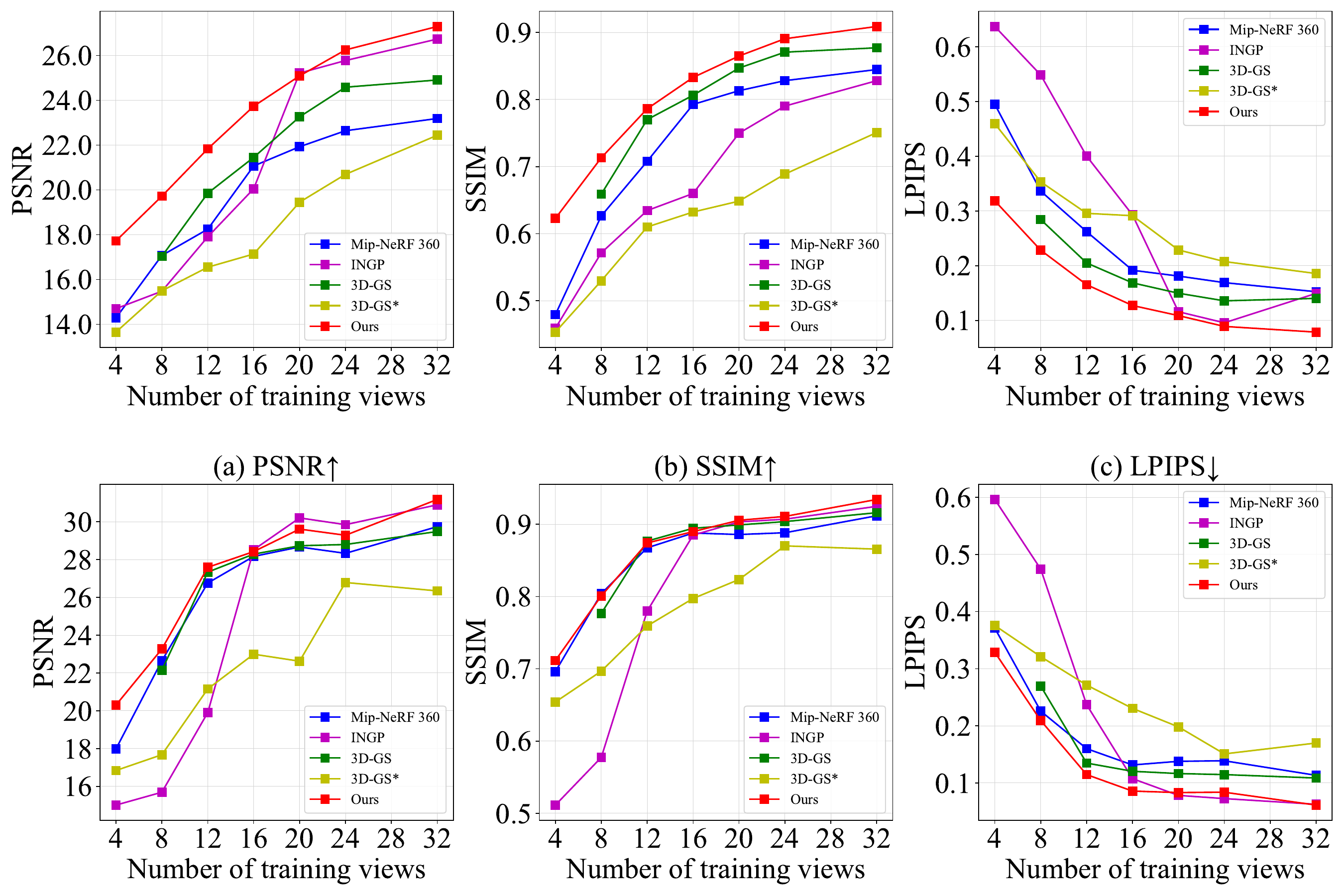}
   \caption{\textbf{Impact of varying training views.} We report results for two scenes, with each row corresponding to the results for one scene. Our method is robust to varying training views and achieves superior quantitative results. }
   \label{fig:views}
\end{figure}
\subsection{Discussion}
\textit{Robustness to the number of training images.}
In Fig.~\ref{fig:views}, we present the variation curve of quantitative results for two scenes under different numbers of training views.
With an increasing number of training views, all method exhibits gradual performance improvement, converging to optimal points. Nevertheless, 3D-GS and INGP struggle to handle inadequate training views, leading to diminished performance with 4-view and 8-view inputs. Our method demonstrates robustness against the number of training views. This can be attributed to the effectiveness of room layout priors, which provide valuable information when inputs are sparse. Additionally, our method consistently outperforms others across the majority of configurations.

\textit{Limitation. }
Despite achieving state-of-the-art performance in panoramic rendering, our method has some limitations. We rely on off-the-shelf networks to obtain layouts and depth priors, which may not yield accurate priors for complex scenes. This concern could be partially mitigated with a more powerful network or the use of a depth camera. Another limitation is that our initialization point cloud occupies more on-disk space, as it is sampled from the dense planes of room layouts. Balancing storage costs and rendering quality may require a meticulously crafted sampling strategy.

\section{Conclusion}
We present a novel layout-guided panoramic Gaussian splatting pipeline named 360-GS, which enables direct panoramic rendering and is robust to sparse inputs. The cornerstone of 360-GS is our $360^{\circ}$ Gaussian splatting algorithm and the incorporation of room layout priors. The $360^{\circ}$ Gaussian splatting algorithm tackles the challenge of modeling projection on the spherical surface by utilizing a perspective projection and mapping, thereby enabling the direct optimization of 3D Gaussians with equirectangular images.
We leverage room layout priors within panoramas during the initialization of 3D Gaussians, providing a more accessible and robust alternative to the SfM point cloud. We additionally introduce a layout-guided regularization to mitigate floater issues and preserve the geometric structure of the room layout. 360-GS supports real-time roaming and delivers state-of-the-art performance on real-world scenes for novel view synthesis.


\bibliographystyle{ACM-Reference-Format}
\bibliography{sample-base}










\end{document}